\documentclass[runningheads,pagebackref,breaklinks,colorlinks]{llncs}

\usepackage{eccv}

\usepackage{eccvabbrv}

\usepackage{graphicx}
\usepackage{booktabs}

\usepackage[accsupp]{axessibility}  %

\usepackage{hyperref}

\usepackage{orcidlink}

\definecolor{cvprblue}{rgb}{0.21,0.49,0.74}
\usepackage{algorithm}
\usepackage{algorithmicx}
\usepackage{algpseudocode}
\usepackage{tikz,xcolor}
\usepackage{multirow}
\usepackage{wrapfig}

\begin{document}

\title{SSAM: Singular Subspace Alignment for Merging Multimodal Large Language Models}

\titlerunning{SSAM: Singular Subspace Alignment for Merging MLLMs}

\author{
    Md Kaykobad Reza\textsuperscript{1}, Ameya Patil\textsuperscript{2}, Edward Ayrapetian\textsuperscript{2}, M. Salman Asif\textsuperscript{1}
}

\institute{
\textsuperscript{1}University of California Riverside~~~~
\textsuperscript{2}Amazon}

\authorrunning{Md Kaykobad Reza et al.}

\maketitle

\begin{abstract}
Multimodal large language models (MLLMs) achieve strong performance by jointly processing inputs from multiple modalities, such as vision, audio, and language. However, building such models or extending them to new modalities often requires large paired datasets and substantial computational resources. Since many pretrained MLLMs (\eg, vision–language or audio–language) are publicly available, we ask whether we can merge them into a single MLLM that can handle multiple modalities? Merging MLLMs with different input modalities remains challenging, partly because of differences in the learned representations and interference between their parameter spaces. To address these challenges, we propose \textbf{Singular Subspace Alignment and Merging (SSAM)}, a \emph{training-free model merging} framework that unifies independently trained specialist MLLMs into a single model capable of handling any combination of input modalities. SSAM maintains modality-specific parameter updates separately and identifies a shared low-rank subspace for language-related parameter updates, aligns them within this subspace, and merges them to preserve complementary knowledge while minimizing parameter interference. Without using any multimodal training data, SSAM achieves state-of-the-art performance across four datasets, surpassing prior training-free merging methods and even jointly trained multimodal models. These results demonstrate that aligning models in parameter space provides a scalable and resource-efficient alternative to conventional joint multimodal training.
\end{abstract}

\section{Introduction}
\label{sec:introduction}

Multimodal Large Language Models (MLLMs) \cite{yin2024mllm-survey, wu2023mllm-survey2} have demonstrated remarkable capabilities across diverse domains by integrating vision, audio, and language understanding within a unified reasoning framework. Building on the success of large language models (LLMs) \cite{zhao2023llmsurvey, minaee2024llmsurvey2}, early multimodal systems aligned different input modalities to a shared language representation space using modality-specific encoders paired with pretrained LLM decoders. This paradigm has proven effective for pairs of modalities such as vision-language \cite{alayrac2022flamingo, liu2023llava, bai2023Qwen-VL}, audio-language \cite{radford2023whisper, gong2024ltu}, video-language \cite{lin2024video-llava, weng2024longvlm}, and point cloud-language \cite{xu2024pointllm, tang2024minigpt}. By leveraging large-scale paired multimodal datasets, these specialized MLLMs achieve strong alignment, generalization, and performance across a wide range of multimodal tasks and modality combinations. 

\begin{wrapfigure}{r}{0.52\textwidth}
  \centering
    \includegraphics[width=0.50\textwidth]{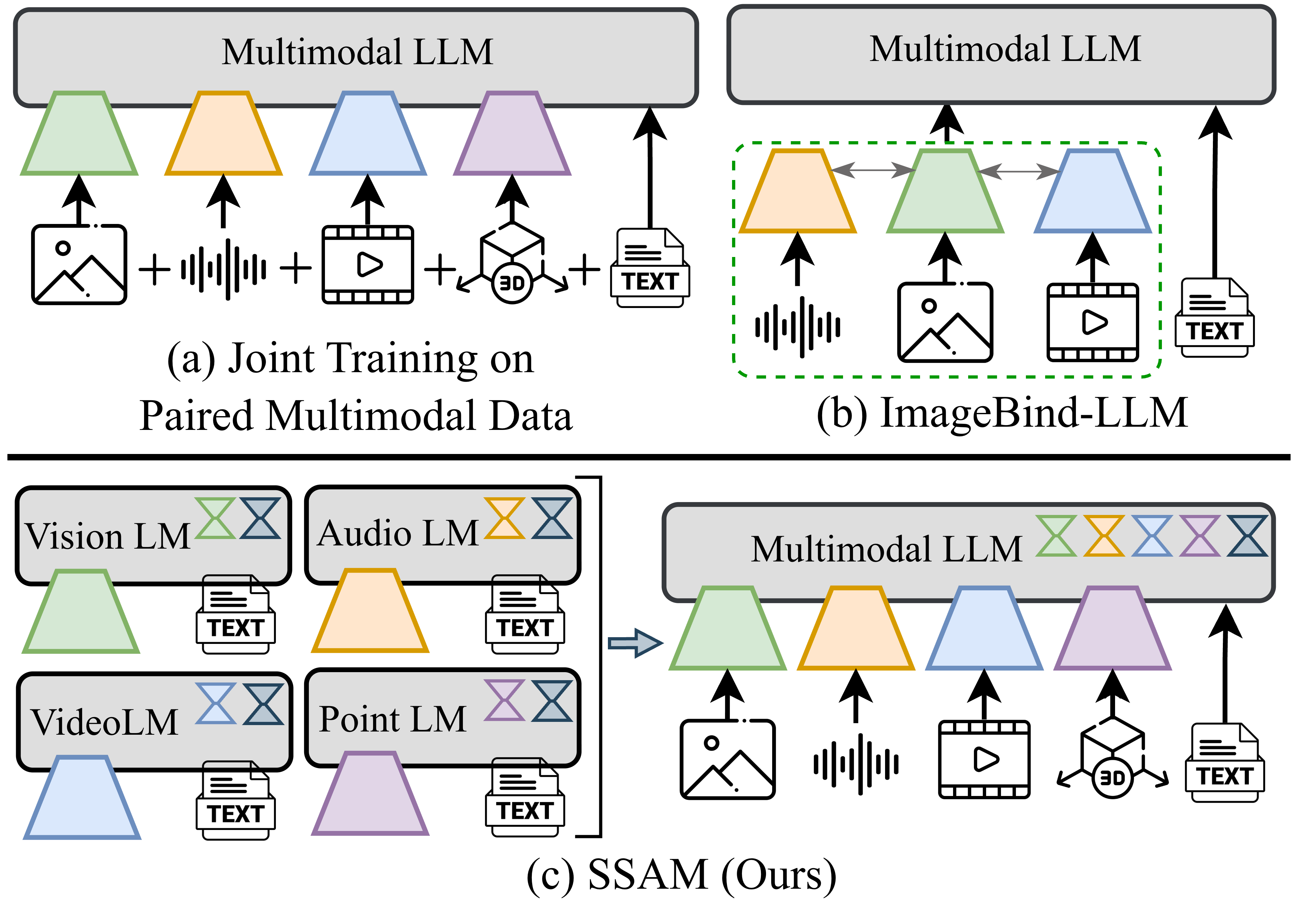}
    \caption{Comparison of multimodal learning paradigms. (a) Conventional methods train a single MLLM on paired multimodal datasets. (b) ImageBind-LLM \cite{han2023imagebind-llm} aligns all modalities to shared image embeddings. (c) SSAM offers a training-free method to merge independently trained specialist MLLMs into a single, unified model via subspace projection.}
   \label{fig:model-merging-overview}
\end{wrapfigure}

Current efforts in this area are geared toward constructing MLLMs that can process multiple modalities simultaneously. One approach incorporates modality-specific \cite{wu2024nextgpt, zhao2023chatbridge} or shared \cite{han2024onellm} encoders with a common language decoder and fine-tunes the model on large-scale paired multimodal datasets, as illustrated in \Cref{fig:model-merging-overview}(a). Training such models or scaling existing MLLMs (\eg, vision–language or audio–language) to new modalities\textbf{ requires massive paired multimodal pretraining and instruction-tuning datasets}, along with significant computational resources.
Alternative methods \cite{han2023imagebind-llm, su2023pandagpt} partially relax data requirements by aligning all modalities to an intermediate image embedding space before connecting to the LLM, as shown in \Cref{fig:model-merging-overview}(b). While this reduces the need for all-modality paired data, it still depends on high-quality image and other modality pairs for alignment. These approaches face two major limitations: first, collecting and synchronizing heterogeneous sensory data (\eg, audio–video–text–point cloud) is costly and often impractical; and second, retraining or extending MLLMs for each new modality remains compute-intensive and laborious.

In this paper, we focus on the challenging problem of combining existing pretrained MLLMs with different input modalities (\eg, vision–language, audio–language) into a single model capable of processing all modalities simultaneously \textbf{without} training on any paired multimodal data, as illustrated in \Cref{fig:model-merging-overview}(c).
Our work is inspired by recent progress in training-free model merging, which demonstrate that knowledge from multiple expert models can be integrated without access to training data. Early studies \cite{ilharco2023taskarithmatic, gargiulo2025tsv, cheng2025wudi, yadav2023ties} revealed that fine-tuned models encode task-specific knowledge as directional ``task vectors'' in parameter space, enabling vector operations to combine model capabilities. \emph{These methods primarily target homogeneous settings (e.g., multiple vision–language, vision only, or language only models).} 
Recent works attempt to bridge this gap via parameter decoupling \cite{chen2024damc} or structured optimization \cite{wei2025optmerge} to reduce task interference. 

\begin{figure}[t]
  \centering

   \includegraphics[width=0.95\linewidth]{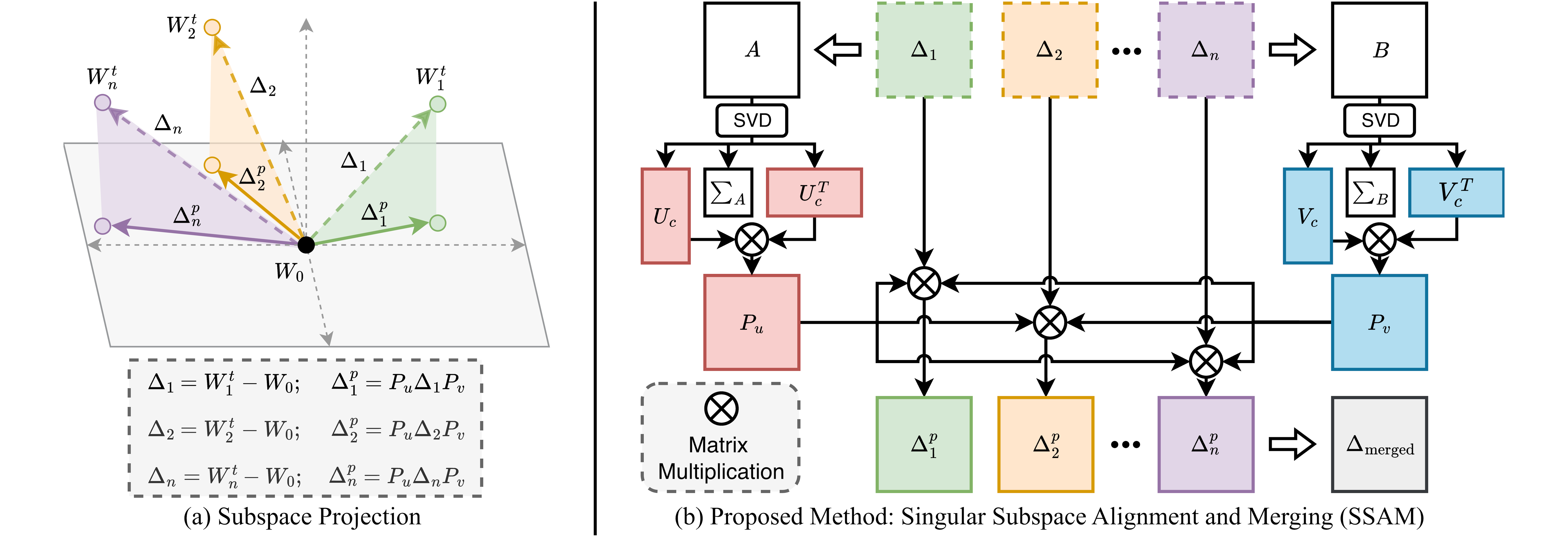}

   \caption{Overview of the proposed SSAM method. (a) Subspace projection: each language vector $\Delta_i$ is projected onto a shared low-rank subspace prior to merging. (b) SSAM constructs covariance matrices $A$ and $B$ from all $\Delta_i$, extracts the shared left and right bases $U_c$ and $V_c$, projects language vectors, and merge them into $\Delta_{\text{merged}}$.}
   \label{fig:ssam-method}
\end{figure}

To address these challenges, we propose \textbf{Singular Subspace Alignment and Merging (SSAM)}, a training-free framework that merges independently trained specialist MLLMs into a single model \emph{capable of processing arbitrary combinations of input modalities}. SSAM is motivated by the observation that task vectors of specialist models are inherently low rank \cite{gargiulo2025tsv} and often span partially overlapping subspaces. Because we consider heterogeneous input modalities, SSAM maintains modality-specific parameter updates separately to reduce modality interference, while computing language-related parameter updates, referred to as \emph{language vectors}, for each specialist model prior to alignment. Unlike prior methods that rely on direct weight interpolation \cite{wei2025optmerge, chen2024damc}, SSAM explicitly identifies a \emph{shared low-rank consensus subspace} that captures consistent parameter update directions across all the specialist MLLMs. Each language vector is projected onto this subspace before merging, ensuring alignment along dominant low-rank directions while filtering out noisy or conflicting updates as shown in \Cref{fig:ssam-method}. This subspace-based alignment mitigates interference in shared language parameters and preserves complementary knowledge from the specialist MLLMs, enabling coherent multimodal reasoning without retraining or access to paired training data (\Cref{fig:qualitative-results}). 

\noindent\textbf{Our main contributions are summarized as follows:}
\begin{itemize}
    \item We propose \textbf{Singular Subspace Alignment and Merging (SSAM)}, a training-free framework that aligns and merges specialist MLLMs by projecting their language vectors onto a shared low-rank subspace, effectively reducing parameter interference while preserving complementary knowledge.
    \item SSAM achieves state-of-the-art performance across four multimodal datasets, consistently outperforming both jointly trained multimodal models and existing training-free model merging approaches (\Cref{sec:results-avqa}, and \Cref{sec:results-mcub}).
    \item We further evaluate SSAM on three domain-specific benchmarks, showing that it preserves specialist capabilities and often improves performance. Extensive ablation studies analyze the impact of subspace rank and merging strategies, demonstrating that subspace alignment enhances stability and cross-modal generalization (\Cref{sec:ablation-studies}).

    \item We present qualitative results demonstrating that the merged model effectively identifies shared semantic cues across image, audio, video, and point cloud modalities, validating SSAM’s coherent multimodal reasoning (\Cref{fig:qualitative-results} and \Cref{sec:supp-qualitative-result}).
\end{itemize}

\section{Related Work}
\label{sec:related-work}

\noindent \textbf{Multimodal Large Language Models (MLLMs).}
Building on the success of large language models (LLMs)~\cite{zhao2023llmsurvey, minaee2024llmsurvey2}, MLLMs~\cite{yin2024mllm-survey, wu2023mllm-survey2} extend pretrained LLMs to process diverse input modalities such as image, audio, and point cloud by aligning them to the language embedding space. Vision–language models such as Flamingo~\cite{alayrac2022flamingo}, LLaVA~\cite{liu2023llava}, and Qwen-VL~\cite{bai2023Qwen-VL} connect vision encoders to LLM decoders via learned projectors, while audio–language models like Whisper~\cite{radford2023whisper} and LTU~\cite{gong2024ltu} apply similar strategies for audio understanding tasks. This paradigm has since been extended to additional modalities including video~\cite{maaz2024video, zhang2023video}, sensor data~\cite{li2025sensorllm, imran2024llasa}, and point clouds~\cite{xu2024pointllm, tang2024minigpt}. Recent efforts aim to develop unified MLLMs capable of processing multiple modalities simultaneously. OneLLM~\cite{han2024onellm} employs a shared encoder across all modalities, whereas NExT-GPT~\cite{wu2024nextgpt} uses modality-specific encoders with projector layers. These models are typically fine-tuned on large-scale paired multimodal datasets for downstream reasoning tasks. Alignment-based frameworks such as ImageBind-LLM~\cite{han2023imagebind-llm} and PandaGPT~\cite{su2023pandagpt} bind all modalities to a common image embedding space following ImageBind~\cite{girdhar2023imagebind} approach before aligning to the language space, reducing but not eliminating the need for paired multimodal data. Despite their success, these methods face two key limitations: collecting large-scale aligned multimodal data is difficult, costly and time consuming due to modality heterogeneity, and training such models requires substantial compute resources.

\noindent \textbf{Model Merging.}
Model merging has emerged as an efficient approach for combining complementary capabilities from multiple expert models. Task Arithmetic~\cite{ilharco2023taskarithmatic} introduced the concept of a \emph{task vector}, representing a directional update in weight space, and showed that arithmetic operations on task vectors can steer model behavior. Subsequent works extend this idea through both training-free and unsupervised test-time adaptation strategies. Training-free approaches merge models without additional training data: Model Soup~\cite{wortsman2022modelsoup} averages model weights, while Fisher Merging~\cite{matena2022fishermerging} applies Fisher-weighted averaging. However, direct averaging often introduces task interference and degrades model performance \cite{gargiulo2025tsv}. To mitigate this, TIES-Merging~\cite{yadav2023ties} prunes and aligns weight signs before merging, DARE~\cite{yu2024dare} randomly drops and rescales parameters, Consensus Merging~\cite{wang2024concensusmerging} removes conflicting updates, WUDI~\cite{cheng2025wudi} explicitly minimizes interference via gradient descent, and Core Space Merging~\cite{panariello2025corespace} projects LoRA-based~\cite{hu2022lora} parameter updates into a compact latent space before merging. Recent studies further explore subspace geometry and orthogonality to improve merging robustness: DO-Merging~\cite{zheng2025Dodo} decouples magnitude and direction components, OSRM~\cite{zhang-zhou-2025-OSRM} enforces orthogonal subspaces to reduce interference, KnOTS~\cite{stoica2024knots} aligns LoRA weights via joint SVD transformation,  VisionFuse~\cite{chen2024visionfuse} combine multiple vision encoders, and Isotropic-Merging~\cite{marczak2025iso-cts} flattens task spectra to enhance alignment. Test-time adaptation methods such as AdaMMS~\cite{du2025adamms} and Representation Surgery~\cite{yang2024rep-surgery} further refine merged models using unlabeled test data. Although these methods show strong results, they are primarily designed for unimodal (vision or language) or homogeneous multimodal settings (only vision–language models) and often require test data for further model finetuning, limiting their applicability to truly multimodal model merging.

\noindent \textbf{Merging MLLMs with Heterogeneous Modalities.}
Recent studies have begun exploring model merging for MLLMs with heterogeneous input modalities (\eg, vision–language and audio–language). DAMC~\cite{chen2024damc} mitigates task interference by processing each modality through separate modality-specific adapters. However, it merges the shared parameters via weighted averaging, which can still introduce interference due to conflicting updates from different source models.
OptMerge~\cite{wei2025optmerge} optimizes task vectors in low-rank space to reduce noise and interference, but direct weight interpolation still struggles to capture complementary information across distinct modalities, often leading to degraded performance.

In contrast, SSAM adopts a training-free approach that preserves modality-specific parameter updates while identifying a shared low-rank subspace for language-specific updates. By capturing the dominant and mutually consistent representational directions across the constituent specialist MLLMs, this subspace alignment strategy mitigates parameter interference during merging. 
SSAM consistently outperforms existing merging methods (\Cref{sec:results-avqa,sec:results-mcub}) and constituent specialist models (\Cref{sec:ablation-studies}) across modality combinations, effectively capturing cross-modal semantics (\Cref{fig:qualitative-results}) without additional fine-tuning.

\section{Proposed Method}
\label{sec:method}
We aim to construct a single unified MLLM, \(M_\text{merged}\), capable of processing arbitrary combinations of input modalities by merging a collection of independently fine-tuned \emph{specialist} models \(\{M_1, M_2, \dots, M_n\}\), each trained on different modality–language pairs (\eg, vision–language, audio–language). Each \(M_i\) consists of two primary components: (1) a modality-specific encoder \(E_i\) that map raw modality inputs (\eg, image, audio, video, or point cloud) into the shared language embedding space, and (2) a language decoder \(W_i\) that jointly processes the multimodal embeddings and textual queries to generate responses. 

Our objective is to derive \(M_\text{merged}\) 
that inherits the multimodal reasoning capabilities of all \(\{M_i\}_{i=1}^n\). \emph{Moreover, we also seek to enable the model to understand arbitrary combinations of input modalities without requiring additional fine-tuning on large-scale paired multimodal data.} For instance, by merging a vision-language model and an audio-language model, the resulting merged model \(M_\text{merged}\) should not only retain the ability to process vision-language and audio-language inputs independently but also gain the capability to reason over combined vision-audio-language inputs. To achieve this, we propose a training-free merging framework: \textbf{Singular Subspace Alignment and Merging (SSAM)}. The key motivation behind SSAM is to preserve modality-specific parameter updates in isolation while identifying a shared, low-rank subspace for the language-specific parameter updates. This subspace is designed to capture the dominant and mutually consistent representational directions across all \(\{M_i\}_{i=1}^n\), thereby mitigating interference during the merging process. A pseudocode for this method is shown in \Cref{alg:ssam} in \Cref{sec:supp-ssam-algorithm}.

\subsection{Problem Formulation}
\label{subsec:objective}
We assume all \(\{M_i\}_{i=1}^n\) share the same language decoder architecture that is fine-tuned from a common pretrained language decoder with weights \(W_0\). \(W_i\) denotes the language decoder weights of the \(i\)-th fine-tuned specialist (\(i \in \{1, 2, \ldots, n\}\)) each fine-tuned to process a separate modality \(m \in \{m_1, m_2, \dots, m_n\}\) paired with language \(t\). We follow a parameter decoupling strategy similar to \cite{chen2024damc} and decouple modality-specific and language-specific parameters as \(W_i^m\) and \(W_i^t\), respectively. An MLLM that is fine-tuned on modality \(m\) paired with language \(t\), we can write the input to any layer \(l\) of the language decoder as \(h_l = [h_l^m, h_l^t]\) where \(h_l^m\) and \(h_l^t\) denote the modality-specific and language-specific hidden states, respectively. For simplicity of notation, we omit the layer subscript \(l\) in the subsequent derivations. 

We calculate the query \(Q\), key \(K\) and value \(V\) matrices for any layer as:
\begin{equation}
    Q = [h^m W_{i, Q}^m, h^t W_{i, Q}^t]; ~~
    K = [h^m W_{i, K}^m, h^t W_{i, K}^t]; ~~
    V = [h^m W_{i, V}^m, h^t W_{i, V}^t];
\end{equation}
where \(W_{i, (.)}^m\) and \(W_{i, (.)}^t\) denote modality-specific and language-specific weights. Attention is applied to the whole sequence and the output is calculated as:
\begin{align}
    [h_a^m, h_a^t] &= \text{Split}(\text{Attention}(Q, K, V) = \text{Split}(\text{softmax}(\frac{QK^\top}{\sqrt{d_k}})V)\\
    [h_o^m, h_o^t] &= [h_a^m W_{i, O}^m, h_a^t W_{i, O}^t]; \quad h_{l+1} = [\text{FFN}_m(h_o^m), \text{FFN}_t(h_o^t)]
\end{align}
where \(d_k\) is the dimension of \(K\) and \(\text{FFN}_m\) and \(\text{FFN}_t\) denote modality-specific and language-specific feed-forward layers. 
During attention, hidden states from all modalities attend to one another through a shared attention map, enabling cross-modal interactions in the joint sequence space. This design allows modality and language tokens to exchange information at every layer while preserving modality-specific parameters, thereby retaining modality-specific representations alongside shared multimodal reasoning.
For our merging method, we keep all the modality-specific parameter updates \(W_{i, (.)}^m\) separate and merge only language-specific weights \(W_{i, (.)}^t\). For simplicity, we denote them as \(W_i^t\).

\subsection{Language Vectors}
\label{subsec:language-vector}
Following prior work~\cite{ilharco2023taskarithmatic, cheng2025wudi}, we define the language-specific parameter updates, referred to as \emph{language vectors}, as
\begin{equation}
\Delta_i = W_i^t - W_0,
\label{eq:task_vector}
\end{equation}
where \(\Delta_i \in \mathbb{R}^{d_{\text{out}} \times d_{\text{in}}}\) represents the parameter shift that adapts the general-purpose language decoder \(W_0\) to a specific multimodal task (e.g., vision–language). We compute these language vectors independently for each linear layer \(l\) of the model. For models adapted using Low-Rank Adaptation (LoRA)~\cite{hu2022lora}, the language vector naturally takes the form \(\Delta_i = A_i^t B_i^t\), where \(A_i^t \in \mathbb{R}^{d_{\text{out}} \times r}\) and \(B_i^t \in \mathbb{R}^{r \times d_{\text{in}}}\) are low-rank matrices.  

Our goal is to estimate a merged language vector \(\Delta_{\text{merged}}\) such that the resulting language decoder  
retains and integrates the capabilities of all \(\{M_i\}_{i=1}^n\).  
Since we only merge the language-specific parameter updates, we hypothesize that the subspaces spanned by the language vectors exhibit significant overlap, making it possible to identify a shared common subspace.
Naïvely 
averaging 
language vectors in the full parameter space (\(\frac{1}{n}\sum_{i=1}^n \Delta_i\)) or in the LoRA space (\(\frac{1}{n}\sum_{i=1}^n A_i^t\) and \(\frac{1}{n}\sum_{i=1}^n B_i^t\))  
often leads to suboptimal performance due to parameter interference (\Cref{sec:ablation-studies}) because conflicting parameter updates in \(\Delta_i, A_i^t, \text{or} B_i^t\) can cancel out~\cite{yadav2023ties, cheng2025wudi}.  
SSAM addresses this by identifying a \emph{shared consensus subspace} that captures consistent update directions across all \(\{\Delta_i\}_{i=1}^n\) and merging the projected language vectors within that subspace.

\subsection{Shared Consensus Subspace Construction}
\label{subsec:subspace}

To identify a shared subspace that best captures the consistent parameter updates across all \(\{\Delta_i\}_{i=1}^n\), we analyze their principal directions of variance.  
Each \(\Delta_i\) represents a directional update in the model’s parameter space, reflecting the adaptation of \(W_0\) toward a specific modality pair.  
Although these updates differ across \(\{W_i\}_{i=1}^n\), they often exhibit correlated directions because they all modify language-related parameters. Our objective is to extract this common structure while suppressing conflicting updates that may introduce noise or interference.

We first compute the row- and column-wise inner-product matrices, \(A\) and \(B\), by aggregating the contributions of all \(\{\Delta_i\}_{i=1}^n\):
\begin{equation}
A = \sum_{i=1}^{n} \Delta_i \Delta_i^\top, 
\qquad
B = \sum_{i=1}^{n} \Delta_i^\top \Delta_i.
\label{eq:covariance}
\end{equation}
These matrices capture the dominant column- and row-space structures shared among the language vectors.  
We then perform Singular Value Decomposition (SVD) as 
$A = U \Sigma_A U^\top, B = V \Sigma_B V^\top$, where \(U\) and \(V\) contain the orthonormal bases corresponding to the principal directions of variation across all \(\{\Delta_i\}_{i=1}^n\).  
The largest singular values in \(\Sigma_A\) and \(\Sigma_B\) correspond to the most consistent and energy-preserving update directions shared among the specialists models.  
By selecting the top-\(k\) eigenvectors in \(U\) and \(V\), we define the orthonormal bases of the \textbf{shared consensus subspace}:
\begin{equation}
U_c = U_{[:,1:k]}, 
\qquad 
V_c = V_{[:,1:k]}.
\label{eq:bases}
\end{equation}

The rank \(k\) controls the effective dimensionality of the shared subspace.  
In practice, \(k\) is chosen to retain the majority of cumulative spectral energy while removing 
task-specific noise~\cite{gargiulo2025tsv, wei2025optmerge}. 
We evaluate \(k \in \{64, 128, 256, 384, 512\}\) and find \(k=128\) empirically provides the best trade-off between shared variance preservation and parameter interference suppression (\Cref{sec:ablation-studies}). 

\noindent \textbf{Justification for shared subspaces.} One might intuitively consider computing the shared subspace by directly applying Truncated SVD to the na\"ively summed language vectors $\bar{\Delta} = \sum_{i=1}^n \Delta_i$. However, the left singular vectors of this direct sum are derived from the covariance matrix:
\begin{equation}
    \bar{\Delta} \bar{\Delta}^\top = \sum_{i=1}^n \Delta_i \Delta_i^\top + \sum_{i \neq j} \Delta_i \Delta_j^\top.
\end{equation}
In the context of MLLMs independently fine-tuned on heterogeneous modalities, the cross-covariance terms ($\sum_{i \neq j} \Delta_i \Delta_j^\top$) often contain parameter interference and spurious cross-task noise. By explicitly defining $A = \sum_{i=1}^n \Delta_i \Delta_i^\top$ and $B = \sum_{i=1}^n \Delta_i^\top \Delta_i$ in \Cref{eq:covariance}, SSAM deliberately drops these noisy cross-terms. Consequently, our bases $U_c$ and $V_c$ minimize the joint column- and row-space projection errors across all \(\{\Delta_i\}_{i=1}^n\):
\begin{align}
    U_c &= \arg\min_{U^\top U = I_k} \sum_{i=1}^n \|\Delta_i - U U^\top \Delta_i\|_F^2,\\ 
    V_c &= \arg\min_{V^\top V = I_k} \sum_{i=1}^n \|\Delta_i - \Delta_i V V^\top\|_F^2.
\end{align}

Next, we define the projection operators for the left and right subspaces and project each \(\Delta_i\) onto the shared consensus subspace as:
\begin{gather}
P_u = U_c U_c^\top \in \mathbb{R}^{d_{\text{out}} \times d_{\text{out}}}, 
\qquad
P_v = V_c V_c^\top \in \mathbb{R}^{d_{\text{in}} \times d_{\text{in}}}
\label{eq:projection_operators}\\
\Delta_i^p = P_u \, \Delta_i \, P_v \in \mathbb{R}^{d_{\text{out}} \times d_{\text{in}}}.
\label{eq:projection}
\end{gather}
This operation aligns the language vectors onto a common coordinate system spanned by the shared bases \(U_c\) and \(V_c\), ensuring that only the most salient update directions are preserved.  
By projecting all \(\{\Delta_i\}_{i=1}^n\) onto this subspace, SSAM captures the shared representational geometry across \(\{M_i\}_{i=1}^n\) while suppressing conflicting updates and reducing parameter interference during merging.

\subsection{Language Vector Merging}
\label{subsec:merging}

After projection, we compute the merged language vector as \(\Delta_{\text{merged}} = \lambda \sum_{i=1}^{n} \Delta_i^p\)
where \(\lambda\) is a scaling coefficient. For simplicity, we can set \(\lambda = \frac{1}{n}\) when the norms of \(\{\Delta_i\}_{i=1}^n\) are similar. However, since each \(M_i\) is fine-tuned independently on different modality pairs, scaling the merged language vector can improve performance on a given downstream task. In practice, \(\lambda\) can be determined using a small validation set for the target task or selected based on the performance of the merged model on individual modality-language pairs~\cite{chen2024damc, ilharco2023taskarithmatic}.

For models fine-tuned using LoRA, we can represent the merged update as \(\Delta_{\text{merged}} \equiv A_{\text{merged}}^t B_{\text{merged}}^t\), where $A_{\text{merged}}^t = U_c$ and $B_{\text{merged}}^t = U_c^\top \lambda \sum_{i=1}^n \Delta_i P_v$. 
This step preserves LoRA’s parameter efficiency and enables direct substitution into pre-trained model checkpoints.  
Bias and normalization parameters are merged via simple averaging.

In summary, SSAM constructs a shared consensus subspace that captures mutually consistent language-specific parameter updates across models, projects them onto this subspace, merges them, and re-factorizes the result into low-rank LoRA form when applicable.
The resulting language decoder effectively integrates knowledge from all specialist MLLMs and enables robust multimodal reasoning without requiring any paired training data or additional fine-tuning.

\section{Experiments and Results}
\label{sec:experiments-and-results}

\subsection{Datasets and Benchmarks}
We evaluate SSAM on four multimodal datasets from two representative benchmarks. In addition, we assess performance on three domain-specific benchmarks to verify that merging preserves the capabilities of the specialist models. A brief overview is provided below, with full dataset details in \Cref{sec:supp-datasets}. We denote video, image, audio, point cloud, and language as V, I, A, P, and L, respectively. 

\noindent \textbf{Audio-Visual Question Answering (AVQA) Benchmarks.}  
The AVQA task evaluates a model’s ability to answer questions on image, audio, and video inputs. We assess SSAM on two large-scale datasets: MUSIC-AVQA \cite{li2022music-avqa} and AVQA  \cite{yang2022avqa}, each comprising three modalities (I, A, and V) paired with language. \textbf{MUSIC-AVQA} focuses on fine-grained multimodal understanding and spatio-temporal reasoning in musical performance. \textbf{AVQA} targets complex interactions and relationships between real-world objects, sounds, and daily activities.

\noindent \textbf{Multimodal Commonality Understanding Benchmark (MCUB).}  
The MCUB benchmark~\cite{chen2024damc} measures a model’s ability to identify shared semantic concepts across diverse input modalities. It includes two variants: \textbf{MCUB-3}, which includes four three-modality combinations (I+V+P, A+V+P, I+A+P, and I+A+V) paired with language, and \textbf{MCUB-4}, which integrates all four modalities (I+A+V+P) paired with language.

\noindent \textbf{Domain-Specific Benchmarks.} To assess whether the merged model preserves the capabilities of the original specialist models, we evaluate it on three domain-specific benchmarks: MMLU \cite{hendryckstest2021mmlu} for language understanding, OCRBench \cite{Liu_2024ocrbench} for vision–language reasoning, and MMAU \cite{sakshi2024mmau} for audio–language tasks. These evaluations demonstrate that merging preserves and, in some cases, improves the performance of the model.

\begin{table}[t]
  \scriptsize
  \centering
  \caption{Performance comparison on MUSIC-AVQA \cite{li2022music-avqa} and AVQA \cite{yang2022avqa} datasets. SSAM achieves the best average performance across all modality combinations and consistently outperforms both training-free model merging baselines and models finetuned on paired multimodal data. 
  }

  \label{tab:avqa_benchmark}
  
  \resizebox{\textwidth}{!}{%
  
  \setlength{\tabcolsep}{8pt}
  \begin{tabular}{l|l|ccccc}
    \toprule
    \textbf{Dataset} & \textbf{Method} & \textbf{V+L} & \textbf{V+I+L} & \textbf{V+A+L} & \textbf{V+I+A+L} & \textbf{Average} \\
    
    \midrule
    
    \multirow{14}{*}{MUSIC-AVQA} & \multicolumn{6}{c}{\textbf{Models finetuned on paired multimodal data}} \\
    
    \cmidrule{2-7}
    
    & ChatBridge-13B \cite{zhao2023chatbridge} & - & - & 43.00 & - & 43.00 \\
    & OneLLM-7B \cite{han2024onellm} & - & - & 47.60 & - & 47.60 \\
    & ImageBind-LLM \cite{han2023imagebind-llm} & 37.24 & 38.76 & 39.72 & 38.16 & 38.47 \\
    & X-InstructBLIP \cite{panagopoulou2023xinstruction-blip} & 45.83 & 41.23 & 48.34 & 47.39 & 45.70 \\
    & Proj-Only \cite{chen2024damc} & 44.93 & 46.64 & 46.17 & 50.21 & 46.99 \\
    
    \cmidrule{2-7}
    
    & \multicolumn{6}{c}{\textbf{Training-free model merging methods}} \\
    \cmidrule{2-7}
    
    & Task Arithmetic \cite{ilharco2023taskarithmatic} & 53.53 & 53.22 & 53.62 & 55.98 & 54.09 \\
    & TSV \cite{gargiulo2025tsv} & 53.81 & 54.26 & 53.63 & 54.73 & 54.11 \\
    & WUDI \cite{cheng2025wudi} & 53.89 & 53.78 & 54.10 & 52.43 & 53.55 \\
    & OptMerge \cite{wei2025optmerge} & 51.83 & 52.17 & 51.05 & 53.17 & 52.06 \\
    & NaiveMC \cite{chen2024damc} & 49.00 & 52.52 & 50.66 & 53.63 & 51.45 \\
    & DAMC \cite{chen2024damc} & 49.09 & 53.08 & 50.91 & \textbf{57.32} & 52.60 \\
    & \textbf{SSAM (Ours)} & \textbf{54.34} & \textbf{54.55} & \textbf{54.68} & 56.30 & \textbf{54.97} \\
    
    \midrule
    \midrule
    
    \multirow{12}{*}{AVQA} & \multicolumn{6}{c}{\textbf{Models finetuned on paired multimodal data}} \\
    \cmidrule{2-7}
    
    & ImageBind-LLM \cite{han2023imagebind-llm} & 51.77 & 51.65 & 55.00 & 54.26 & 53.17 \\
    & X-InstructBLIP \cite{panagopoulou2023xinstruction-blip} & 41.91 & 40.42 & 44.29 & 44.23 & 42.71 \\
    & Proj-Only \cite{chen2024damc} & 67.99 & 66.65 & 67.65 & 66.85 & 67.29 \\
    
    \cmidrule{2-7}
    & \multicolumn{6}{c}{\textbf{Training-free model merging methods}} \\
    \cmidrule{2-7}
    
    & Task Arithmetic \cite{ilharco2023taskarithmatic} & 61.96 & 73.60 & 77.93 & 77.29 & 72.70 \\
    & TSV \cite{gargiulo2025tsv} & 77.88 & 80.28 & 81.91 & 81.28 & 80.34 \\
    & WUDI \cite{cheng2025wudi} & 64.30 & 76.09 & 79.06 & 78.69 & 74.54 \\
    & OptMerge \cite{wei2025optmerge} & 78.37 & 80.26 & 82.10 & 81.52 & 80.56 \\
    & NaiveMC \cite{chen2024damc} & 79.37 & 79.74 & 79.82 & 80.70 & 79.91 \\
    & DAMC \cite{chen2024damc} & 79.15 & 80.30 & 80.40 & 81.31 & 80.29 \\
    & \textbf{SSAM (Ours)} & \textbf{80.45} & \textbf{80.70} & \textbf{82.12} & \textbf{81.87} & \textbf{81.29} \\
    \bottomrule
  \end{tabular}%
  } %
\end{table}

\begin{table*}[t]
    \scriptsize 
    \centering
    \caption{Performance comparison on MCUB benchmark \cite{chen2024damc}. 
    SSAM achieves the best average performance across all modality combinations and consistently outperforms both training-free merging baselines and models finetuned on paired multimodal data.
    }
    \resizebox{\textwidth}{!}{
    \setlength{\tabcolsep}{6pt}
    \begin{tabular}{l|ccccc|c}
    \toprule
    \multirow{2}{*}{\textbf{Method}} & \multicolumn{5}{c|}{\textbf{MCUB-3}} & \textbf{MCUB-4} \\
    & 
    \textbf{I+V+P+L} & 
    \textbf{A+V+P+L} & 
    \textbf{I+A+P+L} & 
    \textbf{I+A+V+L} & 
    \textbf{Average} &
    \textbf{I+A+V+P+L} \\ 
    \midrule
    \midrule
    \multicolumn{7}{c}{\textbf{Models finetuned on paired multimodal data}} \\
    \midrule
    ImageBind-LLM \cite{han2023imagebind-llm} & 31.80 & 31.40 & 33.40 & 35.20 & 32.95 & 32.93 \\
    X-InstructBLIP \cite{panagopoulou2023xinstruction-blip} & 29.40 & 25.20 & 21.20 & 41.40 & 29.30 & 27.94 \\
    Proj-Only \cite{chen2024damc} & 42.60 & 43.80 & 42.80 & 47.40 & 44.15 & 43.00 \\
    \midrule
    \multicolumn{7}{c}{\textbf{Training-free model merging methods}} \\
    \midrule
    Task Arithmetic \cite{ilharco2023taskarithmatic} & 36.80 & 37.60 & 43.80 & 45.00 & 40.80 & 40.53 \\
    TSV \cite{gargiulo2025tsv} & 48.00 & 48.00 & 58.00 & 51.60 & 51.40 & 52.29 \\
    WUDI \cite{cheng2025wudi} & 38.20 & 40.40 & 44.40 & 44.40 & 41.85 & 40.95 \\
    OptMerge \cite{wei2025optmerge} & 52.80 & 53.60 & 58.60 & 53.80 & 54.70 & 55.88 \\
    NaiveMC \cite{chen2024damc} & 53.00 & 51.00 & 58.80 & 56.00 & 54.70 & 54.03 \\
    DAMC \cite{chen2024damc} & 58.20 & 58.80 & \textbf{65.60} & \textbf{56.60} & 59.80 & 60.08 \\
    \textbf{SSAM (Ours)} & \textbf{59.40} & \textbf{61.00} & 65.40 & 55.60 & \textbf{60.35} & \textbf{62.27} \\
    \bottomrule
    \end{tabular}
    }
    \label{tab:mcub_results}
\end{table*}

\subsection{Implementation Details}
\label{sec:implementation-details}

We consider four input modalities: Image (I), Audio (A), Video (V), and Point Cloud (P) paired with Language (L). For all experiments, we use the pretrained specialist model checkpoints provided by \cite{chen2024damc}, which include vision–language, audio–language, video–language, and point–language models. Each specialist uses Vicuna-7B-v1.5 \cite{zheng2023vicuna} as language decoder, CLIP-ViT-L-336px \cite{radford2021clip} as image encoder, BEATs-Iter3+ \cite{chen2022beats} as audio encoder, LanguageBind \cite{zhu2023languagebind} video encoder, and PointLLM \cite{xu2024pointllm} point encoder. The audio encoder employs a Q-Former \cite{li2023blip} with 32 query tokens as the projector, while other modalities use MLP projectors.

During merging, we experiment with ranks $k \in \{64, 128, 256, 384, 512\}$ and find that $k=128$ yields overall better performance (\Cref{sec:ablation-studies}). 
We report results from published works whenever available. For training-free model merging methods~\cite{ilharco2023taskarithmatic, gargiulo2025tsv, cheng2025wudi, wei2025optmerge}, we reproduce results using the same checkpoints with the public codebase of \cite{wei2025optmerge}. Additional implementation details are provided in supplementary \Cref{sec:supp-implementation-details}.

\subsection{Results on AVQA Benchmarks}
\label{sec:results-avqa}

We summarize the performance of SSAM on the two AVQA datasets in \Cref{tab:avqa_benchmark}. We compare SSAM with two categories of methods: (1) jointly trained multimodal models that rely on large paired multimodal datasets, including ChatBridge-13B~\cite{zhao2023chatbridge}, OneLLM~\cite{han2024onellm}, ImageBind-LLM~\cite{han2023imagebind-llm}, X-InstructBLIP~\cite{panagopoulou2023xinstruction-blip}, and Proj-Only~\cite{chen2024damc}; and (2) training-free model merging approaches such as Task Arithmetic~\cite{ilharco2023taskarithmatic}, TSV~\cite{gargiulo2025tsv}, WUDI~\cite{cheng2025wudi}, OptMerge~\cite{wei2025optmerge}, NaiveMC~\cite{chen2024damc}, and DAMC~\cite{chen2024damc}.

\subsubsection{Performance on MUSIC-AVQA dataset}
For MUSIC-AVQA, training-free merging methods generally outperform models trained on paired multimodal data, highlighting the effectiveness of model merging without retraining. SSAM achieves the highest average accuracy of 54.97\%, surpassing the best finetuned baseline, Proj-Only (46.99\%), by 7.98 percentage points. 
Within the training-free merging category, SSAM attains the best average accuracy and maintains consistent performance across all modality combinations. It achieves the top scores in three subsets: V+L (54.34\%), V+I+L (54.55\%), and V+A+L (54.68\%). In V+I+A+L setting, DAMC slightly exceeds SSAM (57.32\% vs. 56.30\%), but SSAM achieves the best overall mean accuracy across all combinations. These results indicate that SSAM effectively integrates complementary information from different specialist models while mitigating interference between modalities. 

\subsubsection{Performance on AVQA dataset}
On the AVQA dataset, SSAM achieves an average accuracy of 81.29\%, outperforming both finetuned and training-free baselines. Compared to the best finetuned model, Proj-Only (67.29\%), SSAM improves by 14 percentage points. Among training-free merging methods, SSAM consistently achieves the highest accuracy across all modality combinations.

\subsection{Results on MCUB Benchmarks}
\label{sec:results-mcub}

The results for MCUB-3 and MCUB-4 datasets are summarized in \Cref{tab:mcub_results}. We compare SSAM with (1) models finetuned on paired multimodal data~\cite{han2023imagebind-llm, panagopoulou2023xinstruction-blip, chen2024damc}, and (2) existing training-free model merging methods~\cite{ilharco2023taskarithmatic, gargiulo2025tsv, cheng2025wudi, wei2025optmerge, chen2024damc}.

\subsubsection{Performance on MCUB-3 dataset}
Compared to models trained on paired multimodal data, training-free merging methods show substantial improvements, underscoring their potential for scalable and  efficient merging of MLLMs. SSAM achieves the best overall performance with an average accuracy of 60.35\%, surpassing the best finetuned baseline, Proj-Only (44.15\%), by 16.20 percentage points. Within the training-free category, SSAM attains the highest average accuracy, outperforming the next-best method, DAMC (59.80\%). A closer analysis reveals complementary strengths: SSAM leads in the I+V+P+L (59.40\%) and A+V+P+L (61.00\%) subsets, while DAMC performs slightly better in I+A+P+L (65.60\%) and I+A+V+L (56.60\%). Despite these minor variations, SSAM maintains strong overall performance across all subsets, demonstrating consistent generalization across diverse modality combinations.

\subsubsection{Performance on MCUB-4 dataset}
MCUB-4 presents a more challenging scenario, requiring reasoning over all four modalities (I, A, V and P) simultaneously. In this setting, SSAM achieves an accuracy of 62.03\%, outperforming all training-free baselines, including DAMC (60.08\%), and surpassing the best jointly trained model, Proj-Only (43.00\%), by 19.03 percentage points. 

\begin{figure}[t]
  \centering
  \includegraphics[width=1.0\linewidth]{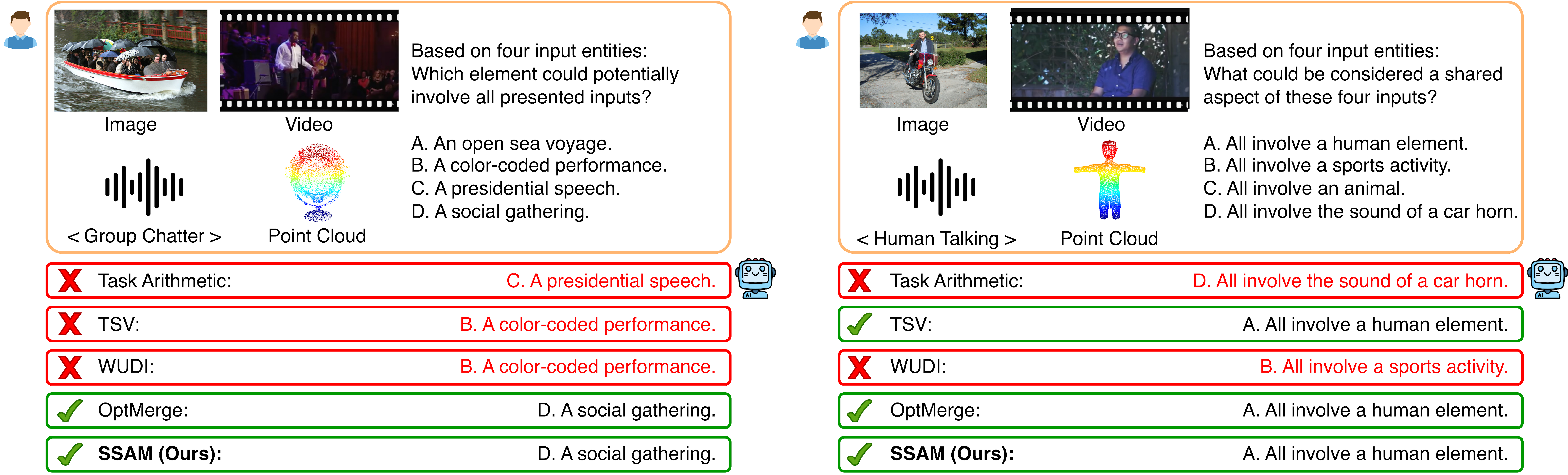}

   \caption{Qualitative examples on commonality understanding on the MCUB-4 dataset.}
   \label{fig:qualitative-results}
\end{figure}

\Cref{fig:qualitative-results} presents qualitative examples from the MCUB-4 dataset, showing that the merged model effectively identifies shared semantic cues across image, audio, video, and point cloud modalities, although none of the specialist models was fine-tuned on such combinations. Additional examples are provided in supplementary \Cref{sec:supp-qualitative-result}. These results highlight SSAM’s ability to integrate complementary modality information and maintain coherent multimodal reasoning across unseen modality combinations.

\begin{table}[t] 
    \scriptsize
    \centering
    
    \begin{minipage}[t]{0.48\textwidth}
        \caption{Comparison of specialist vs. merged MLLMs on the MUSIC-AVQA and AVQA datasets. The merged model consistently outperforms individual specialists across modality pairs and generalizes well to the unseen four-modality (V+I+A+L) setting.}
        \label{tab:specialist-vs-merged-model-performance}
        \setlength{\tabcolsep}{2pt}
        \resizebox{\textwidth}{!}{
            \begin{tabular}{lcccccc}
            \toprule
            \textbf{Dataset} & \textbf{Model} & \textbf{V+L} & \textbf{I+L} & \textbf{A+L} & \textbf{V+I+A+L} & \textbf{Avg.} \\
            \midrule
            \midrule
            MUSIC- & Specialist & 48.88 & 51.22 & 33.25 & -- & 44.45 \\
            AVQA & Merged     & \textbf{54.34} & \textbf{53.46} & \textbf{47.15} & \textbf{56.30} & \textbf{52.81} \\
            \midrule
            \multirow{2}{*}{AVQA} 
            & Specialist & 79.09 & 75.52 & 44.18 & -- & 66.26 \\
            & Merged     & \textbf{80.45} & \textbf{76.33} & \textbf{64.21} & \textbf{81.87} & \textbf{75.72} \\
            \bottomrule
            \end{tabular}
        }
    \end{minipage}%
    \hfill 
    \begin{minipage}[t]{0.48\textwidth}
        \centering
        \caption{Comparison of specialist vs. merged MLLMs across three domain-specific benchmarks. VLM, Vid-LM, and ALM denote Vision-Language, Video-Language, and Audio-Language models, respectively. SSAM (ours) consistently outperforms individual specialists.}
        \setlength{\tabcolsep}{3pt}
        \resizebox{\textwidth}{!}{
        \begin{tabular}{l|ccc|c}
        \toprule
        \multirow{2}{*}{\textbf{Benchmarks}} & \multicolumn{3}{c|}{\textbf{Specialist MLLMs}} & \textbf{Merged} \\
         & \textbf{VLM} & \textbf{Vid-LM} & \textbf{ALM} & \textbf{SSAM} \\
        \midrule
        MMLU      & 50.52 & 50.25 & 42.62 & \textbf{51.82} \\
        OCRBench & 319   & --    & --    & \textbf{329}   \\
        MMAU (test-mini)     & --    & --    & 48.50 & \textbf{58.90} \\
        \bottomrule
        \end{tabular}
        }
        \label{tab:benchmark_comparison}
    \end{minipage}
    
\end{table}

\begin{figure}[t]
  \centering
  \includegraphics[width=1.0\linewidth]{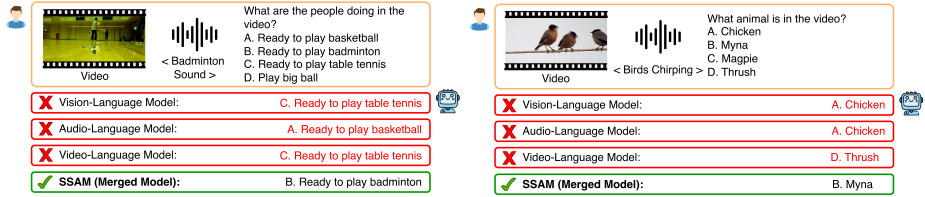}

   \caption{Qualitative examples on the AVQA dataset.}
   \label{fig:qualitative-results-specialist-models}
\end{figure}

\subsection{Ablation Studies}
\label{sec:ablation-studies}

\noindent\textbf{Model Merging Enhances Cross-Modal Performance.}
We compare the merged model with its constituent specialist vision–language, audio–language, and video–language models across different modality–language tasks and summarize the results in \Cref{tab:specialist-vs-merged-model-performance}. The merged model consistently outperforms individual specialists on both MUSIC-AVQA and AVQA datasets, improving average performance by +8.36\% and +9.46\%, respectively. Notably, it generalizes effectively to the unseen four-modality (V+I+A+L) setting, despite none of the specialists being trained on this combination. We also provide qualitative examples in \Cref{fig:qualitative-results-specialist-models}, demonstrating that the merged model yields more accurate predictions than its specialist counterparts. Further examples are available in supplementary \Cref{sec:supp-qualitative-result}. 
These gains arise from the subspace alignment mechanism, which projects language vectors onto a shared low-rank consensus subspace. By emphasizing consistent representational directions and suppressing noisy or conflicting updates, this alignment enables the model to integrate complementary strengths from all the specialist models, resulting in more robust and accurate multimodal reasoning.

\begin{figure}[t]
    \centering
    
    \begin{minipage}[t]{0.54\textwidth}
        \centering
        \includegraphics[width=\linewidth]{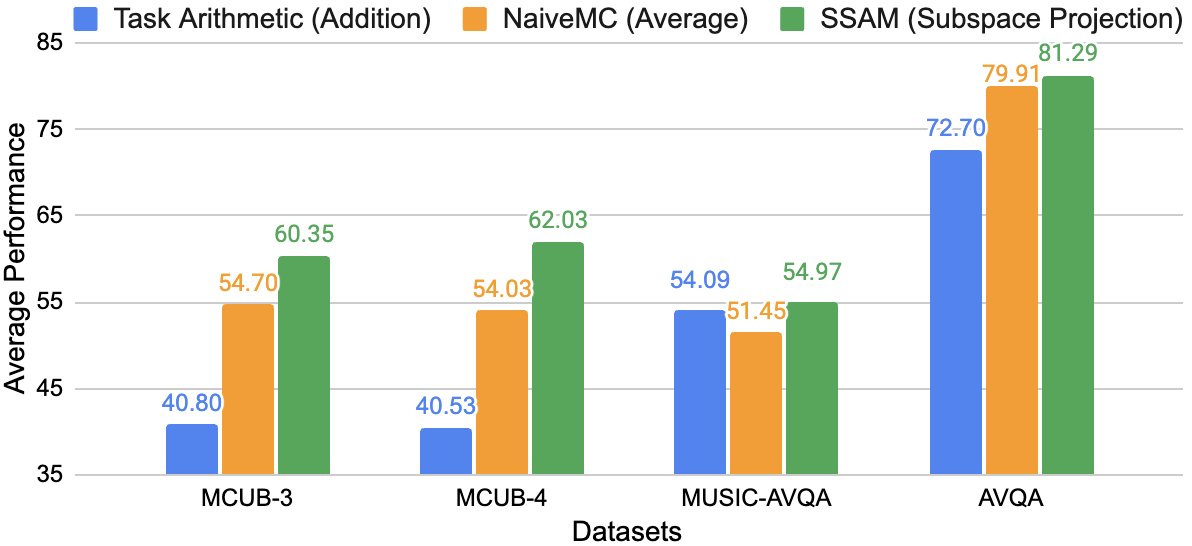}
        \caption{Comparison with simple model-merging baselines.
        SSAM consistently outperforms them across all four datasets.}
        \label{fig:averaging-vs-ssam}
    \end{minipage}%
    \hfill
    \begin{minipage}[t]{0.44\textwidth}
        \centering
        \includegraphics[width=\linewidth]{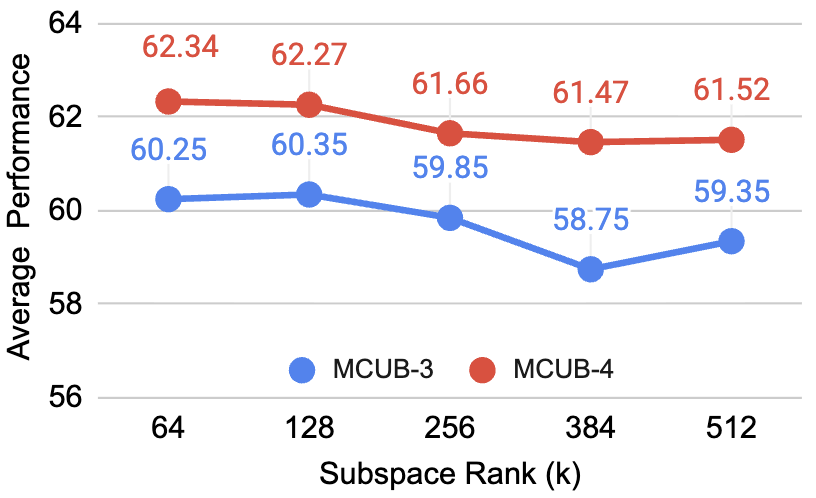}
        \caption{Effect of subspace rank ($k$) on average performance. Empirically $k=128$ offers slightly better performance.}
        \label{fig:rank-vs-performance}
    \end{minipage}
    
\end{figure}

\noindent\textbf{Comparison with Simple Merging Strategies.}  
We compare SSAM against two standard training-free merging baselines: Task Arithmetic~\cite{ilharco2023taskarithmatic}, which directly adds the language vectors, and NaiveMC~\cite{chen2024damc}, which averages them. As shown in Figure~\ref{fig:averaging-vs-ssam}, direct addition or averaging results in suboptimal performance across all datasets (likely due to conflicting parameter updates and parameter interference)~\cite{gargiulo2025tsv, cheng2025wudi, wei2025optmerge}. In contrast, SSAM projects language vectors into a shared low-rank subspace before merging, aligning consistent update directions while filtering modality-specific variations. This projection consistently improves performance across all four datasets, demonstrating that subspace alignment offers a more stable and coherent alternative to simple linear combination methods.

\noindent\textbf{Merging Preserves Specialist Capabilities.}
To assess whether the merged model preserves the capabilities of the original specialist models, we evaluate it on three domain-specific benchmarks. The results are summarized \Cref{tab:benchmark_comparison}. We evaluate SSAM on MMLU \cite{hendryckstest2021mmlu} benchmark to assess potential degradation of core language capabilities. SSAM slightly outperforms all specialist models indicating that merging does not degrade language capabilities of the original specialist models.
To evaluate vision–language and audio-language reasoning, we evaluate SSAM on OCRBench \cite{Liu_2024ocrbench} and MMAU (test-mini) \cite{sakshi2024mmau}, respectively. Results show that SSAM does not cause performance degradation on these benchmarks and slightly outperforms specialist models on both of them. 
These results indicate that SSAM does not degrade performance or suffer from catastrophic forgetting on domain-specific tasks and often provides performance improvements.

\noindent\textbf{Effect of Subspace Rank (\(k\)).}  
To determine the optimal subspace dimensionality, we evaluate SSAM with ranks \(k \in \{64, 128, 256, 384, 512\}\) on the MCUB-3 and MCUB-4 datasets. As shown in Figure~\ref{fig:rank-vs-performance}, ranks 64 and 128 exhibit similar performance, whereas increasing the rank gradually decreases performance. We hypothesize that higher ranks introduce redundant or conflicting updates within the language vectors, leading to parameter interference. Rank \(k=128\) empirically yields overall better performance across both datasets and is adopted in all the experiments. The reported averages are computed over all modality combinations, with detailed breakdowns provided in supplementary \Cref{sec:supp-rank-vs-performance}.

\section{Conclusion}
\label{sec:conclusion}

We propose Singular Subspace Alignment and Merging (SSAM), a training-free framework for merging independently trained MLLMs. Unlike conventional joint multimodal training, which requires large paired datasets and substantial computational resources, SSAM integrates specialist MLLMs into a single model capable of processing arbitrary combinations of input modalities.
SSAM preserves modality-specific parameter updates while identifying a shared low-rank subspace that captures dominant and mutually consistent directions across language-specific parameters. By aligning and merging updates within this subspace, SSAM constructs a unified MLLM that retains modality-specific representations while capturing shared knowledge in a common language space. This design mitigates cross-modal interference and enables effective integration without paired multimodal training data or additional fine-tuning.
Extensive experiments on four datasets show that SSAM achieves state-of-the-art performance, outperforming prior training-free merging methods and models trained on large-scale paired multimodal data. Evaluation on three additional benchmarks shows that SSAM preserves and often improves performance on domain-specific tasks. Overall, SSAM demonstrates that alignment in parameter space can serve as a practical alternative to costly joint multimodal training while maintaining strong performance.

\section*{Acknowledgment}
This work is supported in part by NSF awards 2046293 and 2406199 and Amazon Gift awards.

\bibliographystyle{splncs04}
\bibliography{main}

\newcommand{
    \beginsupplement}{%
        \setcounter{table}{0}
        \renewcommand{\thetable}{S\arabic{table}}%
        \setcounter{figure}{0}
        \renewcommand{\thefigure}{S\arabic{figure}}%
        \setcounter{section}{0}
        \renewcommand{\thesection}{S\arabic{section}}%
    }

\beginsupplement

\clearpage
\setcounter{page}{1}

\begin{center}
    {\Large \bf SSAM: Singular Subspace Alignment for Merging Multimodal Large Language Models}\\
    \vspace{8pt}
    {\Large Supplementary Material}
\end{center}

\section{Datasets and Benchmarks}
\label{sec:supp-datasets}

We evaluate the performance of our merged model on four multimodal datasets from two representative benchmarks. In addition, we also assess performance on three domain-specific benchmarks to verify that merging preserves the capabilities of the specialist models. Detailed descriptions of the datasets and benchmarks are provided below.

\noindent\textbf{MUSIC-AVQA.}
The MUSIC-AVQA dataset~\cite{li2022music-avqa} is a large-scale audio–visual question answering benchmark constructed from musical performance videos. It contains 45,867 question–answer pairs derived from 9,288 videos, totaling over 150 hours of content. The dataset emphasizes fine-grained multimodal understanding and spatio-temporal reasoning, covering 9 question types and 33 question templates. Each question requires joint reasoning over different combinations of image, audio, and video modalities paired with a language query. 

\noindent\textbf{AVQA.}
The AVQA dataset~\cite{yang2022avqa} is a large-scale audio–visual question answering benchmark designed to evaluate multimodal understanding of real-world objects, events, and activities. It comprises 57,015 videos depicting diverse daily scenarios and 57,335 carefully constructed question–answer pairs. The questions are designed to require joint reasoning over image, audio, and video modalities, capturing temporal dependencies, causal interactions, and cross-modal relationships. Each question may rely on different combinations of the three input modalities paired with a language query.

\noindent\textbf{Multimodal Commonality Understanding Benchmark (MCUB).}
The MCUB benchmark~\cite{chen2024damc} evaluates the ability of a model to infer shared semantic concepts from heterogeneous multimodal inputs. Each sample includes inputs from multiple modalities (image, audio, video, and point cloud) along with a textual query, and the model must select the correct answer from four candidate answers. The benchmark assesses both cross-modal alignment and higher-level conceptual reasoning, making it a challenging benchmark for unified multimodal understanding. MCUB offers two evaluation datasets. 
\textbf{MCUB-3} contains four subsets, each comprising three of the four input modalities paired with language: image–video–point cloud–language, audio–video–point cloud–language, image–audio–point cloud–language, and image–audio–video–language. \textbf{MCUB-4} includes all four input modalities (image, audio, video, and point cloud) paired with language. Together, these settings enable a comprehensive assessment of a model’s ability to generalize across diverse and unseen modality combinations. Additional dataset details, preprocessing steps, and annotation procedures are available in \cite{chen2024damc}.

\noindent \textbf{Domain-Specific Benchmarks.} 
To assess whether the merged model preserves the capabilities of the original specialist models, we evaluate it on three domain-specific benchmarks covering language only, vision–language, and audio–language tasks. Specifically, we use MMLU \cite{hendryckstest2021mmlu}, a widely used benchmark for evaluating general language understanding across diverse academic and professional subjects; OCRBench \cite{Liu_2024ocrbench}, which measures vision–language reasoning with a focus on optical character recognition and text understanding in images; and MMAU \cite{sakshi2024mmau}, a benchmark designed to evaluate multimodal audio–language reasoning and understanding. We report performance on the MMAU (test-mini) subset. These evaluations verify whether the merged model retains the domain expertise of the original specialist models. Our results show that merging preserves, and in some cases slightly improves, performance compared to the original specialist models.

\section{Implementation Details}
\label{sec:supp-implementation-details}

We use the pretrained specialist models released by \cite{chen2024damc}, which include checkpoints for vision–language, audio–language, video–language, and point cloud–language models. Each specialist model is composed of a modality-specific encoder, a projection module, and a Vicuna-7B-v1.5 \cite{zheng2023vicuna} language decoder. Below, we summarize the encoder and projector architectures for each modality:

\begin{itemize}
    \item \textbf{Vision-Language Model:} CLIP-ViT-L-336px \cite{radford2021clip} is used as the vision encoder, followed by a two-layer MLP projector to align image embeddings to the language embedding space.
    
    \item \textbf{Audio-Language Model:} BEATs-Iter3+ \cite{chen2022beats} is used as the audio encoder and a Q-Former \cite{li2023blip} with 32 learnable query tokens serves as the projection module.
    
    \item \textbf{Video-Language Model:} The video encoder from LanguageBind \cite{zhu2023languagebind} is used, coupled with an MLP projector similar to the vision projector.
    
    \item \textbf{Point Cloud-Language Model:} The point cloud encoder from PointLLM \cite{xu2024pointllm} is used with a corresponding MLP projector to align point cloud embeddings to the language embedding space.
\end{itemize}

All specialist models follow a two-stage training procedure as described in \cite{chen2024damc}. In stage 1, only the projectors are trained while keeping the modality-specific encoders and the language decoder completely frozen. In stage 2, the language decoder is fine-tuned using LoRA \cite{hu2022lora} with rank 128 and $\alpha = 256$, while keeping all modality-specific encoders and projectors frozen. We use the publicly released checkpoints from \cite{chen2024damc} directly for all the experiments without any additional fine-tuning or modification. All merging is performed in a training-free setting.

For evaluating baseline merging methods, Task Arithmetic \cite{ilharco2023taskarithmatic}, TSV \cite{gargiulo2025tsv}, WUDI \cite{cheng2025wudi}, and OptMerge \cite{wei2025optmerge}, we apply their public implementations to the same specialist checkpoints to ensure a fair comparison. Results for other baselines are taken from their respective publications.

\begin{algorithm}[t]
    \caption{SSAM Algorithm}
    \label{alg:ssam}
    \textbf{Input:} Pretrained model weight $W_0$; \\
        \hspace*{1cm} Language-specific weights from specialist models $\{W_i^t\}_{i=1}^n$;\\
        \hspace*{1cm} Rank $k$; Number of layers $L$; \\
        \hspace*{1cm} Scaling coefficient $\lambda$.\\
    \textbf{Output:} Merged language-specific weight $W^t_{\text{merged}}$
    \begin{algorithmic}[1]
    
    \State $W^t_{\text{merged}} \gets \{\}$
    \For{$l = 1$ to $L$}
        \Statex \(\triangleright\) {\textbf{ Step 1: Calculate language vectors for  layer $l$} \(\triangleleft\)} \hfill 
        \For{$i = 1$ to $n$}
            \State $\Delta_{i, l} \gets W^t_{i, l} - W_{0, l}$
        \EndFor
        
        \Statex
        \Statex \(\triangleright\)  {\textbf{Step 2: Compute shared subspaces for  layer $l$}} \(\triangleleft\) 
        \State $A_l \gets \sum_{i=1}^n \Delta_{i, l}\Delta_{i, l}^\top$ 
        \State $B_l \gets \sum_{i=1}^n \Delta_{i, l}^\top\Delta_{i, l}$
        \State $U_l$, $\Sigma_{A, l}$, $U_l^\top \gets \text{SVD}(A_l)$
        \State $U_{c, l} \gets U_{l, [:,1:k]}$ 
        \State $V_l$, $\Sigma_{B, l}$, $V_l^\top \gets \text{SVD}(B_l)$
        \State $V_{c, l} \gets V_{l, [:,1:k]}$
        
        \Statex
        \Statex  \(\triangleright\) {\textbf{Step 3: Subspace projection and merging}} \(\triangleleft\) 
        \State $P_{u, l} \gets U_{c, l} U_{c, l}^\top$ 
        \State $P_{v, l} \gets V_{c, l} V_{c, l}^\top$ %
        \For{$i = 1$ to $n$}
            \State $\Delta_{i, l}^p \gets P_{u, l} \Delta_{i, l} P_{v, l}$
        \EndFor
        \State $\Delta_{\text{merged}, l} \gets \lambda \sum_{i=1}^n \Delta_{i, l}^p$
    
        \Statex
        \Statex  \(\triangleright\)  {\textbf{Step 4: Final merged language-specific weight for layer $l$}} \(\triangleleft\)
        \State $W^t_{\text{merged}}(l) \gets W_{0, l} + \Delta_{\text{merged}, l}$ 
        
    \EndFor
    \State \Return $W^t_{\text{merged}}$
    \end{algorithmic}
\end{algorithm}

\begin{table}[ht]
    \scriptsize
    \centering
    \caption{Performance of SSAM with different subspace ranks ($k$) on the MCUB-3 and MCUB-4 datasets. I, A, V, P, and L denote Image, Audio, Video, Point Cloud, and Language, respectively. Rank $k = 128$ empirically yields overall better performance.}
    \setlength{\tabcolsep}{6pt}
    \resizebox{\textwidth}{!}{%
        \begin{tabular}{c|ccccc|c}
            \toprule
            \textbf{Subspace} & \multicolumn{5}{c|}{\textbf{MCUB-3}} & \textbf{MCUB-4} \\ 
            \cmidrule{2-7}
             \textbf{Rank (k)} & \textbf{I+V+P+L} & \textbf{A+V+P+L} & \textbf{I+A+P+L} & \textbf{I+A+V+L} & \textbf{Average} & \textbf{I+A+V+P+L} \\
             \midrule
             \midrule
            64 & 58.20 & \textbf{61.40} & 65.00 & \textbf{56.40} & 60.25 & \textbf{62.34} \\
            128 & \textbf{59.40} & 61.00 & \textbf{65.40} & 55.60 & \textbf{60.35} & 62.27 \\
            256 & 59.20 & 59.20 & 64.80 & 56.20 & 59.85 & 61.66 \\
            384 & 58.40 & 57.80 & 64.40 & 54.40 & 58.75 & 61.47 \\
            512 & 58.60 & 58.20 & 65.20 & 55.40 & 59.35 & 61.52 \\
            \bottomrule
        \end{tabular}
    }
    \label{tab:rank-vs-performance-mcub}
\end{table}

\begin{figure*}[t]
  \centering
  \includegraphics[width=1.0\linewidth]{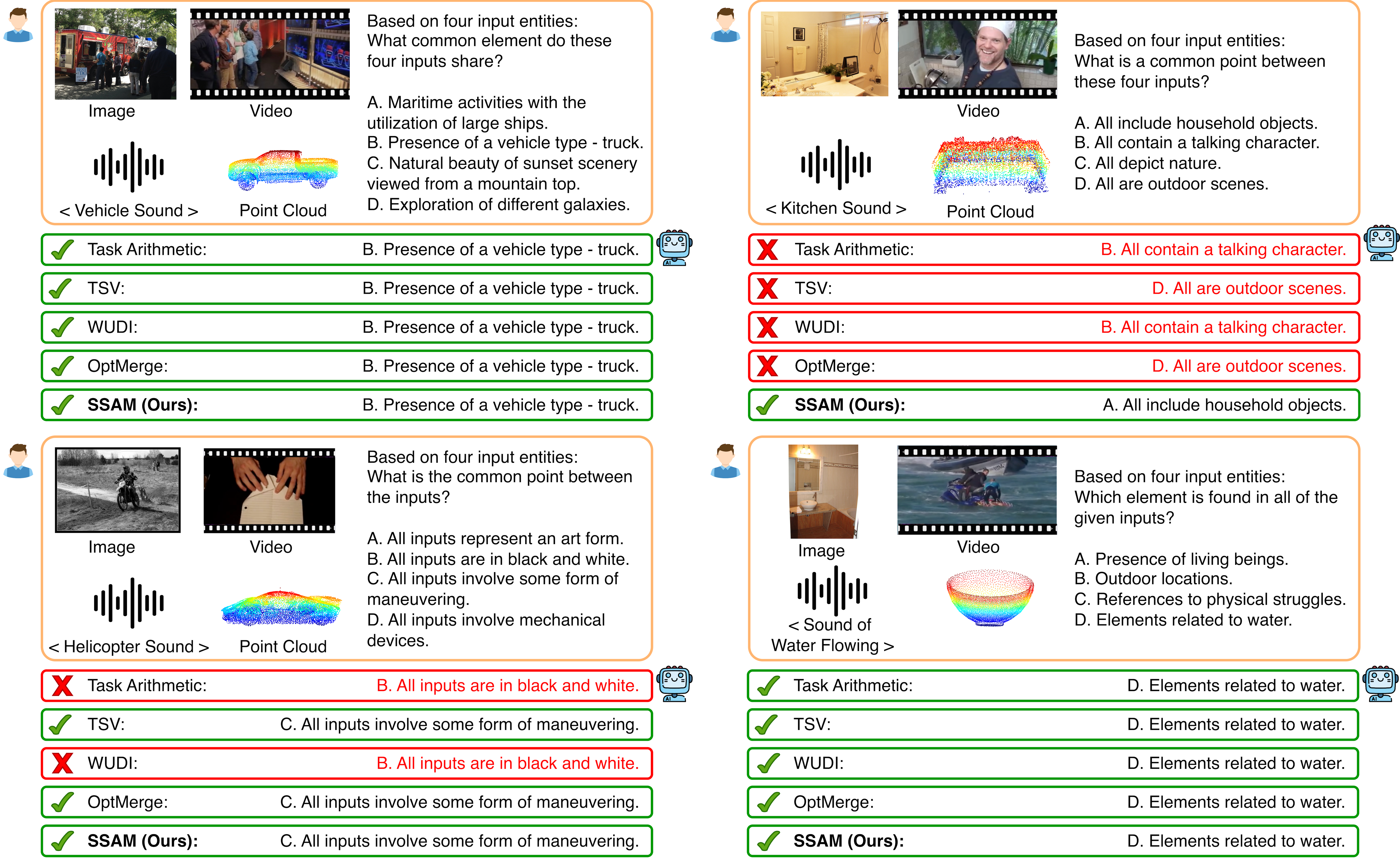}

   \caption{Qualitative results on the MCUB-4 \cite{chen2024damc} dataset with four input modalities (image, video, audio, and point cloud) paired with natural-language queries. SSAM demonstrates strong cross-modal reasoning and provides coherent answers in complex multimodal settings, even though none of the specialist models were trained on this modality combination.}
   \label{fig:supp-qualitative-results-mcub}
\end{figure*}

\begin{figure*}[t]
  \centering
  \includegraphics[width=1.0\linewidth]{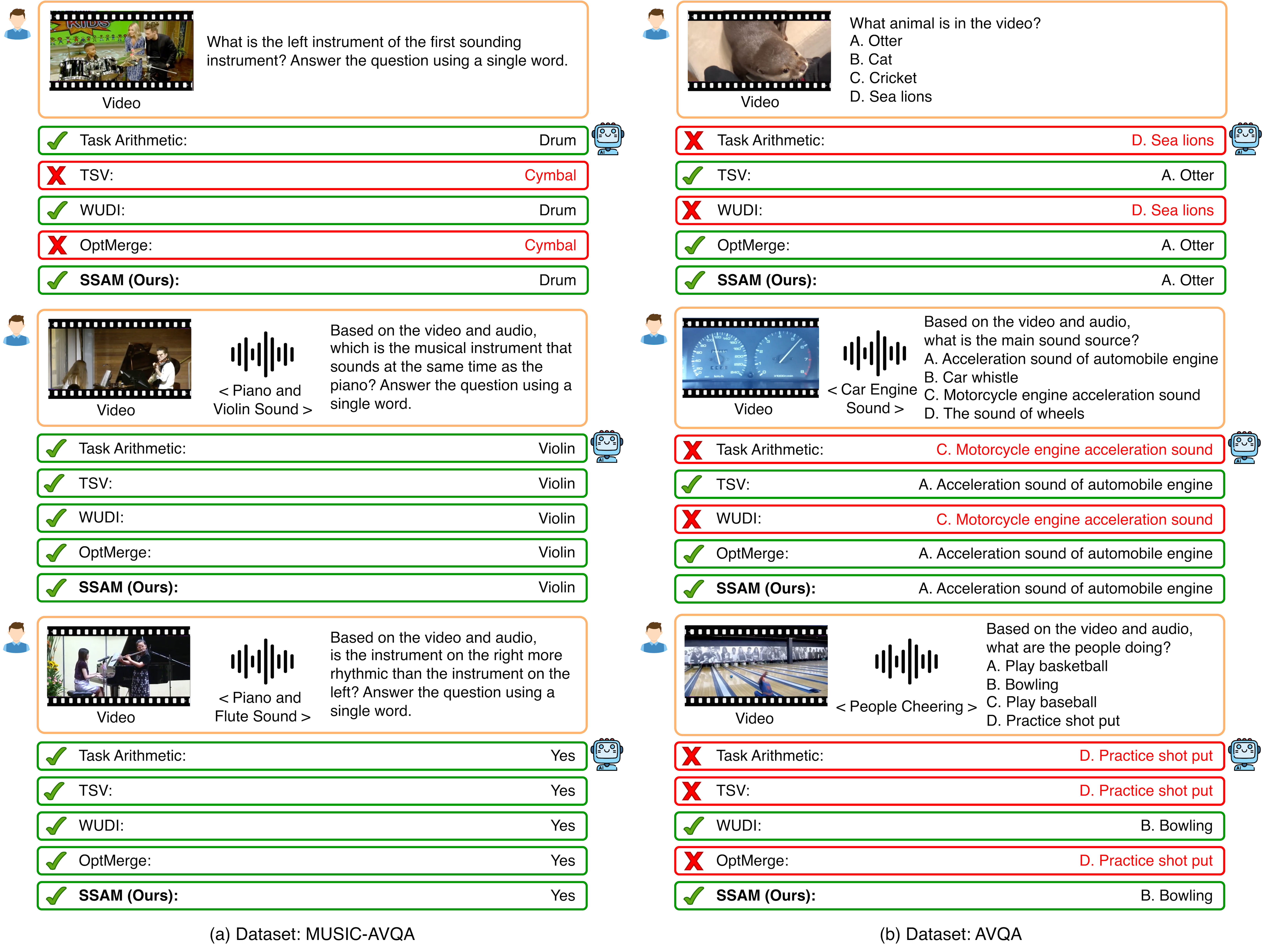}

   \caption{Qualitative results on the MUSIC-AVQA \cite{li2022music-avqa} and AVQA \cite{yang2022avqa} datasets with video and audio as the input modalities paired with natural-language queries. SSAM provides consistent responses, demonstrating stable performance on multimodal queries despite the specialist models not being trained on paired video-audio-language data.}
   \label{fig:supp-qualitative-results-avqa}
\end{figure*}

\begin{figure*}[t]
  \centering
  \includegraphics[width=1.0\linewidth]{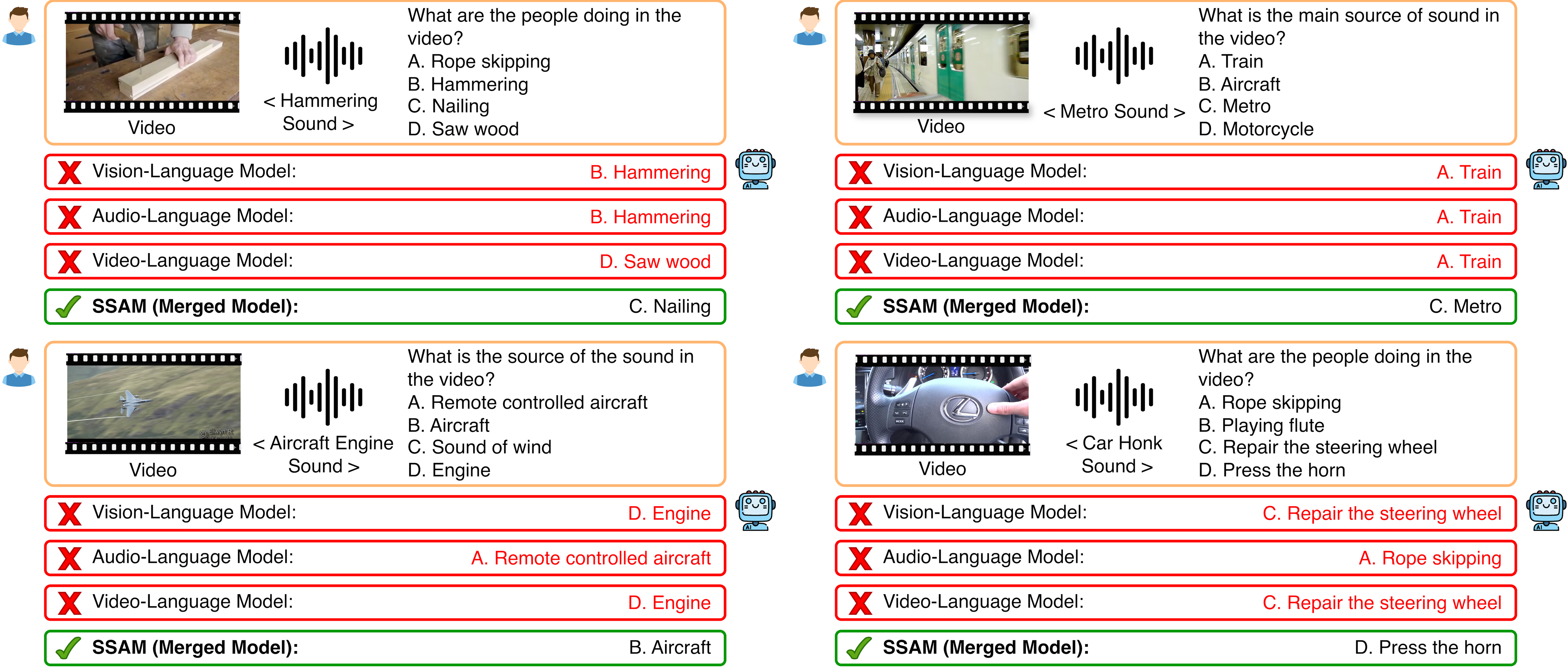}

   \caption{Qualitative results on the AVQA \cite{yang2022avqa} dataset with video and audio as input modalities paired with natural-language queries. SSAM effectively combines information from both audio and video to produce more accurate and consistent responses, demonstrating stable performance on multimodal queries compared to individual specialist models.}
   \label{fig:supp-qualitative-results-avqa-specialist}
\end{figure*}

\section{Proposed SSAM Algorithm}
\label{sec:supp-ssam-algorithm}

The key idea of SSAM is threefold: (i) compute the \emph{language vector} \(\Delta_i\) for each specialist language decoder \(W_i\), representing its fine-tuning direction in parameter space; (ii) identify a shared low-rank subspace that captures dominant and mutually consistent update directions across all language vectors; and (iii) project and merge these language vectors within this subspace to form the final language-specific decoder weight \(W^t_{\text{merged}}\). This process integrates shared knowledge from all specialist models while minimizing parameter interference. As mentioned in \Cref{sec:method}, these language vectors are computed independently for each linear layer \(l\) of the model. 
The pseudocode for SSAM is shown in \Cref{alg:ssam}.

\section{Selecting Optimal Subspace Rank \texorpdfstring{($k$)}{k}}
\label{sec:supp-rank-vs-performance}

We evaluate the performance of the proposed method, SSAM, with subspace ranks \(k \in \{64, 128, 256, 384, 512\}\) on the MCUB-3 and MCUB-4 datasets to determine the optimal subspace dimensionality. In this section, we provide a detailed performance analysis for each modality combination. As shown in Table~\ref{tab:rank-vs-performance-mcub}, both \(k=64\) and \(k=128\) achieve similar performance, with \(k=128\) performing marginally better. Increasing the rank beyond this point gradually decreases the performance of the merged model. We hypothesize that higher ranks introduce redundant or conflicting updates within the language vectors, leading to parameter interference. Therefore, we use \(k=128\) for all experiments.

\section{Qualitative Results}
\label{sec:supp-qualitative-result}

We present additional qualitative examples from the MCUB-4 \cite{chen2024damc} dataset in Figure~\ref{fig:supp-qualitative-results-mcub} and from the MUSIC-AVQA \cite{li2022music-avqa} and AVQA \cite{yang2022avqa} datasets in Figure~\ref{fig:supp-qualitative-results-avqa}. Each example includes outputs from SSAM alongside baseline merging methods: Task Arithmetic~\cite{ilharco2023taskarithmatic}, TSV~\cite{gargiulo2025tsv}, WUDI~\cite{cheng2025wudi}, and OptMerge~\cite{wei2025optmerge}. Across these datasets, SSAM generates coherent responses across heterogeneous input modality combinations, while other baseline merging methods often fail under unseen modality combinations. These examples illustrate that SSAM effectively integrates complementary information from image, audio, video, and point cloud modalities. Notably, SSAM generalizes to unseen modality combinations (\eg, audio–video–language, image–audio–video–point cloud–language), despite none of the specialist models being trained to handle such combinations jointly.

Qualitative examples on the AVQA \cite{yang2022avqa} dataset in \Cref{fig:supp-qualitative-results-avqa-specialist} demonstrate the advantage of SSAM in multimodal reasoning scenarios requiring joint audio–video understanding. We use the first frame from each video as the image modality for the vision–language model. In these examples, individual specialist models (vision–language, audio–language, and video–language) produce incorrect predictions when relying on a single modality. In contrast, the merged model (SSAM) integrates complementary cues from both audio and video to infer the correct answer. For instance, SSAM correctly identifies actions such as \emph{nailing} by combining the visual scene with the characteristic hammering sound, recognizes a \emph{metro} from the combination of subway visuals and audio, distinguishes an \emph{aircraft} by jointly reasoning about the visual appearance and engine sound, and correctly identifies a \emph{car horn} from the steering-wheel interaction in video with the honking audio. These examples highlight SSAM’s ability to capture information across modalities and provide more accurate responses than individual specialist models without being trained on such modality combinations.

Overall, these qualitative results reinforce that the proposed subspace alignment and merging mechanism effectively combines complementary cross-modal information and supports robust multimodal reasoning across diverse and previously unseen modality combinations.

\section{Discussion and Future Directions}
\label{sec:discussion}

\noindent\textbf{Discussion.}  
Model merging approaches such as SSAM offer a practical and scalable way to unify specialist MLLMs without retraining or access to paired multimodal data. As the number of fine-tuned models on open-source platforms like Hugging Face~\cite{wolf2019huggingface} continues to grow, model merging provides an efficient means to reuse, combine, and extend these models. It promotes model reuse, reduces data collection and training costs, preserves privacy by avoiding data
\clearpage
\noindent sharing, and enables distributed development where independently trained specialist models can be integrated into a unified multimodal system. 

\noindent\textbf{Limitations and Future Directions.}  
To maintain fairness with existing benchmarks and account for computational constraints, our experiments focused on models with approximately 7B parameters. We focus on merging models with the same architecture; extending to heterogeneous architectures remains an open challenge. We evaluated SSAM on four datasets that include diverse combinations of image, audio, video, point cloud, and language modalities. However, there is currently no comprehensive benchmark specifically designed to evaluate \emph{MLLM merging}. Developing such benchmarks would enable broader and more systematic evaluation across additional modalities, such as depth, thermal, IMU, or medical imaging, which are essential for assessing generalization to heterogeneous and complex modality interactions. Finally, the safety and alignment of merged MLLMs remain open challenges. Understanding how merging affects factual consistency, ethical behavior, and robustness is important for ensuring reliable and trustworthy multimodal systems.

\end{document}